\documentclass[conference]{IEEEtran}
\IEEEoverridecommandlockouts
\usepackage{cite}
\usepackage{amsmath,amssymb,amsfonts}
\usepackage{algorithmic}
\usepackage{graphicx}
\usepackage{textcomp}
\usepackage{pgfplots}
\pgfplotsset{compat=1.17} 
\usepgfplotslibrary{groupplots}
\usepackage{siunitx}      
\usepackage{xcolor}
\usepackage{booktabs} 
\usepackage{multirow}
\usepackage{siunitx} 
\usepackage{url}
\usepackage{tikz}
\usepackage{comment}
\usepackage[acronym]{glossaries}
\usepackage{tcolorbox}
\makeglossaries
\newacronym{hci}{HCI}{Human--Computer Interaction}
\newacronym{iot}{IoT}{Internet-of-Things}
\newacronym{tinyml}{TinyML}{Tiny Machine Learning}
\newacronym{tof}{ToF}{Time-of-Flight}
\newacronym{ir}{IR}{Infrared}
\newacronym{cnn}{CNN}{Convolutional Neural Network}
\newacronym{mcu}{MCU}{Micrcontroller Unit}
\newacronym{fov}{FoV}{Field of View}
\newacronym{tsne}{t-SNE}{T-distributed Stochastic Neighbor Embedding}
\newacronym{gap}{GAP}{Global Average Pooling}
\newacronym{relu}{ReLU}{Rectified Linear Unit}
\newacronym{bn}{BN}{Batch Normalization}

\usetikzlibrary{positioning, matrix, calc}
\usepackage[caption=false]{subfig}\usepackage{xstring} 
\usepackage{orcidlink}
\newcommand{\drawconfusionmatrix}[1]{
    \begin{tikzpicture}[scale=0.58] 
        \def\labels{Call, Fist, Okay, One, Peace, Stop, Stop Inv}
        
        \foreach \val [count=\i from 0] in {#1} {
            \pgfmathtruncatemacro{\x}{mod(\i, 7)}
            \pgfmathtruncatemacro{\y}{6 - floor(\i / 7)}
            \ifnum\val=0 \def\mycolor{white} \else \def\mycolor{blue!\val!white} \fi
            \fill[\mycolor] (\x, \y) rectangle ++(1, 1);
            \ifnum\val>50 \def\txtcol{white} \def\fontweight{\bfseries} \else \def\txtcol{black} \def\fontweight{} \fi
            \ifnum\val>0
                \node[text=\txtcol, font=\sffamily\tiny\fontweight, align=center] at (\x+0.5, \y+0.5) {\val\%};
            \fi
        }
        
        \foreach \l [count=\i from 0] in \labels {
            \node[anchor=east, font=\sffamily\scriptsize] at (0, 6.5-\i) {\l};
            \node[anchor=east, rotate=45, font=\sffamily\scriptsize] at (\i+0.5, -0.1) {\l};
        }
        
        \draw[gray!50] (0,0) rectangle (7,7);
    \end{tikzpicture}
}

\begin{document}

\title{Efficient Sensor Fusion for Gesture Recognition on Resource-Constrained Devices%
\thanks{This work was carried out in the EssilorLuxottica Smart Eyewear Lab, a Joint Research Center between EssilorLuxottica and Politecnico di Milano}%
}

\author{
\IEEEauthorblockN{
Pietro Bartoli\orcidlink{0009-0001-7705-7916},
Christian Veronesi\orcidlink{0009-0000-7070-8686},
Tommaso Bondini\orcidlink{0009-0006-9818-0768},
Andrea Giudici\orcidlink{0000-0002-5680-5727},
Franco Zappa\orcidlink{0000-0003-1715-501X}
}
\IEEEauthorblockA{
\textit{Dipartimento di Elettronica, Informazione e Bioingegneria (DEIB)} \\
\textit{Politecnico di Milano, Milan, Italy} \\
pietro.bartoli@polimi.it
}
}

\maketitle

\begin{abstract}
Gesture recognition is a cornerstone of Human-Computer Interaction (HCI) for smart eyewear, enabling natural and device-free control in augmented reality environments.
Traditional vision-based approaches face significant challenges regarding power consumption, computational latency, and user privacy. 
This paper proposes a lightweight, privacy-preserving gesture recognition system based on the fusion of low-resolution Time-of-Flight (ToF) and Infrared (IR) thermal sensors. 
We used an $8\times8$ multizone ToF sensor (VL53L8CH) and an $8\times8$ IR array (AMG8833) to capture complementary depth and thermal cues. 
A compact Convolutional Neural Network (CNN) with a specialized grouped-convolution architecture is designed to fuse these modalities efficiently on a microcontroller (MCU). 
Experimental results on a custom dataset of 7 static gestures, validated via k-fold cross-validation, demonstrate that the proposed fusion strategy significantly outperforms single-sensor baselines with an accuracy of 92.3\% and a macro F1-score of 0.93.
Finally, on-device benchmarks on STM32F4 and STM32H7 MCUs confirm the system's suitability for resource-constrained wearables, requiring only 6,343 parameters ($\approx$ \SI{7}{kB}) and achieving millisecond-level inference latency with a total system power of $\approx$\SI{50}{mW}.
\end{abstract}


\section{Introduction}

Gesture recognition has become a core technology in \gls{hci}, enabling natural and touchless device control \cite{Fertl_2025}. 
This need is particularly pronounced in wearables with limited interaction surfaces, most notably smart eyewear, whose compact form factor limits the feasibility of conventional physical inputs \cite{Koutromano_2023,Kim_2019}. 
Broadly, reliable wearable interaction demands navigating the strict memory, compute, and energy constraints of continuous on-body operation.

Ideally, an input system for such platforms must satisfy three challenging requirements: (i) \textbf{Low Power Consumption}, to ensure all-day battery life; (ii) \textbf{Privacy Preservation}, to allow usage in public spaces without capturing personally identifiable information; and (iii) \textbf{Robustness}, to operate reliably under varying environmental conditions \cite{Gallardo_2023,Contoli_2024}.
 
First, the stringent energy budget of battery-powered wearables makes continuous sensing and inference particularly challenging. 
\gls{tinyml} has emerged as a practical approach to run neural models directly on \gls{mcu} within tight memory and power envelopes, enabling always-on interaction without prohibitive battery drain \cite{Capogrosso_2024,Heydari_2025}.

Second, pervasive sensing in wearables and \gls{iot} systems raises significant privacy and "social surveillance" concerns. 
Continuous acquisition of motion, physiological, and environmental signals can reveal sensitive information about users’ health, habits, and social interactions, motivating privacy-preserving mechanisms that span from sensor-level data collection to higher-layer processing \cite{Datta_2018,Zhang_2025,Yang_2024}. 
Among the available sensing modalities, cameras are particularly problematic since always-on video capture can record faces and contextual details making always-on video acquisition particularly problematic for privacy-sensitive interaction design \cite{Opaschi_2020,Bhardwaj_2024,Gallardo_2023}.
Low-resolution, non-textured sensing provides a privacy-preserving alternative to RGB cameras in wearables and \gls{iot} systems. 
Furthermore, its minimal data volume naturally aligns with \gls{tinyml} paradigms, allowing compact convolutional architectures to operate within strict memory and energy budgets.


Third, real-world wearable deployment demands robustness across diverse users and environmental conditions.
Existing non-camera solutions typically rely on a single modality, making them vulnerable to performance degradation when this specific cue becomes ambiguous or noisy \cite{Newaz_2024,Vandersteegen_2020,Safa_2024,Kaseris_2025}.
This is particularly problematic for fine-grained gesture sets where different commands share similar silhouettes or trajectories and are distinguished only by subtle geometric or appearance differences.

To address these challenges, we propose fusing low-resolution \gls{tof} and \gls{ir} sensors. 
\gls{tof} provides low-overhead depth maps, while \gls{ir} offers robust thermal contrast; at \mbox{$8\times 8$} pixels, both modalities limit the capture of identifiable biometric detail, making them well suited to privacy-preserving interaction.
Building on this front-end, we: (i) design a lightweight \gls{cnn} using a \textit{grouped-early-fusion} architecture to process geometric and thermal features independently; (ii) evaluate alternative fusion strategies to validate our design; and (iii) quantify multimodal advantages over single-sensor baselines.


\begin{figure}[!t]
    \centering
    \resizebox{.95\linewidth}{!}{%
        \includegraphics{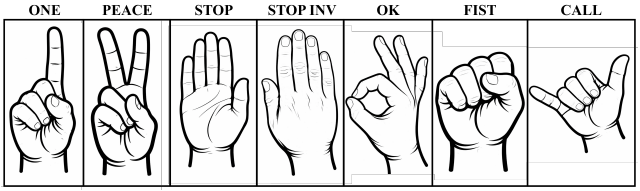}%
    }
    \caption{The seven static hand gestures included in the dataset.}
    \label{fig:gestures}
\end{figure}

\begin{figure}[!t]
    \centering
    \includegraphics[width=.9\linewidth]{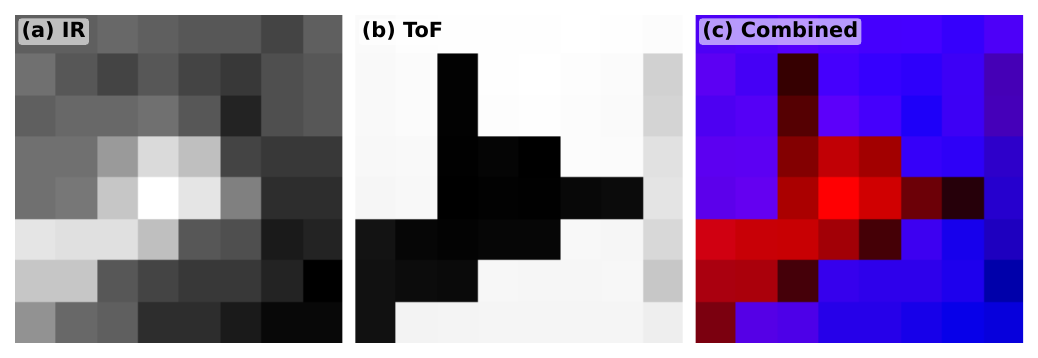}
    \caption{Example of a synchronized multimodal input sample from the dataset. 
    (a) $8\times8$ thermal frame acquired by the AMG8833, 
    (b) corresponding $8\times8$ depth map from the VL53L8CH, and 
    (c) channel-wise fused RGB-encoded representation used as network input.}
    \label{fig:multimodal_sample}
\end{figure}

\begin{figure*}[!t]
    \centering
    \includegraphics[width=.9\linewidth]{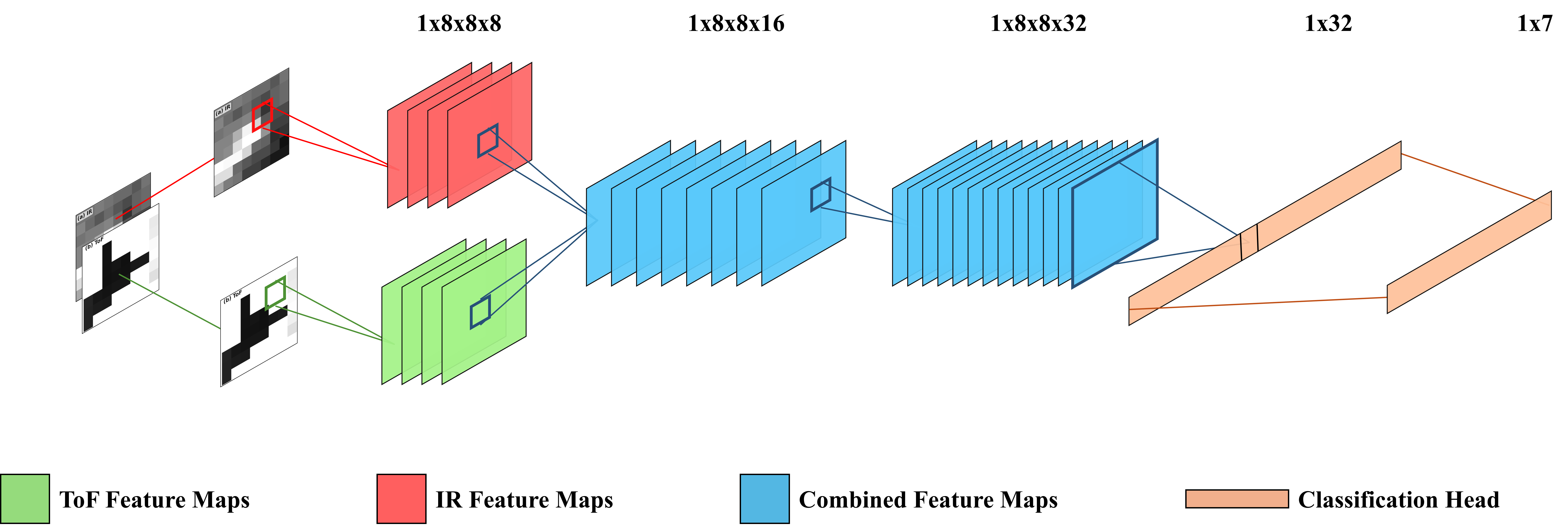}
    \caption{Schematic of the \textbf{Early Fusion} architecture. The number of feature maps corresponds to the filter count for each layer ($8, 16, 32$). Note that in the first layer, the \gls{ir} (red) and \gls{tof} (green) feature maps are visually separated to illustrate the logical independence enforced by grouped convolutions; however, in deployment, they constitute a single contiguous tensor.}
    \label{fig:network}
\end{figure*}

\section{Related Work}
Low-resolution \gls{ir} arrays have been widely explored for privacy-preserving gesture and activity recognition, typically using \mbox{$8\times 8$} or \mbox{$32\times 24$} thermopile sensors (e.g., Panasonic Grid-EYE, Melexis) to capture coarse ``thermal silhouettes'' of the hand or body that remain informative for classification while suppressing identifiable texture details \cite{Vandersteegen_2020,Yin_2021,Chen_2025}.
These sensors have been applied to near-field mid-air interaction, device-free indoor activity recognition, and discrete hand-gesture classification from thermal signatures, and have also been used in multimodal setups where thermal array is fused with ultra-wideband radar for deep-learning-based gesture recognition \cite{Skaria_2020}.
In parallel, sparse multizone \gls{tof} grids drive depth-based interaction, enabling dynamic gesture classification, posture recognition and swipe tracking \cite{Bartoli_2024,Ma_2024,wang_2023}.

\section{Materials and Methods}
\label{sec:materials_methods}


\subsection{Acquisition Setup and Dataset}

A compact multimodal sensing node was developed to acquire a synchronized dataset of static hand gestures for training and evaluating the proposed sensor-fusion pipeline. 
The platform integrates two co-located low-resolution sensors on a shared I$^2$C bus, interfaced to an \textbf{STM32F4 Nucleo} \gls{mcu} to ensure a deterministic, lockstep readout. 
The thermal channel employs a \textbf{Panasonic AMG8833} \gls{ir} sensor array, providing an $8\times 8$ temperature matrix over 0--80~\si{\celsius} with $\pm$2.5~\si{\celsius} accuracy and up to 10~Hz output rate, with a \SI{60}{\degree} horizontal \gls{fov}. 
The depth channel uses an \textbf{STMicroelectronics VL53L8CH} multizone \gls{tof} module configured to output an $8\times 8$ distance map, with a \SI{60}{\degree} diagonal \gls{fov}. 

Data collection was performed through an always-on firmware sampling both modalities synchronously at \SI{5}{Hz}.
Each synchronized thermal/depth pair was timestamped and streamed to a host PC. 
In this configuration, the sensor power budget remains minimal: the AMG8833 (\SI{3.3}{V}) draws \SI{14.9}{mW}, while the VL53L8CH draws \SI{32.3}{mW} total (\SI{3.3}{V}/\SI{1.8}{V} rails) for autonomous $8\times8$ ranging at \SI{5}{Hz} with a \SI{60}{\degree} \gls{fov}.

Using the proposed acquisition pipeline, we collected a synchronized thermal-depth gesture dataset.
Seven static classes were defined to cover \gls{hci} commands (Fig.~\ref{fig:gestures}), \textbf{Call}, \textbf{Fist}, \textbf{Okay}, \textbf{One}, \textbf{Peace}, \textbf{Stop}, and \textbf{Stop Inv}, chosen to mirror common categories in the RGB gesture dataset \textit{HaGRID}~\cite{Nuzhdin_2024}.
The dataset includes \textbf{8,400} paired frames (1,200 samples per class) collected from four participants (two females, two males; age $28 \pm 1$). 
The dataset was balanced across gesture classes and balanced across participants, with both modalities stored as raw $8\times8$ matrices.
Figure~\ref{fig:multimodal_sample} reports an example of the \textit{Call} gesture acquired synchronously by the two sensors and the corresponding fused tensor representation.


\subsection{Experimental Methodology}
\label{sec:methods}

This work first identifies the most effective fusion strategy among the considered architectures and subsequently evaluates whether multimodal fusion provides a consistent advantage over single-modality baselines (\gls{ir}-only and \gls{tof}-only).
Given the limited cohort size, we adopted a sample-wise stratified 5-fold cross-validation protocol to obtain a statistically robust estimate of classification performance while preserving class balance across folds.
Model performance is assessed via classification accuracy (reported as \textbf{mean $\pm$ std} over five folds). 
Qualitative error analysis is conducted on the best-performing fold via confusion matrices, while architectural feasibility on resource-constrained \glspl{mcu} is evaluated through parameter count.


Networks are trained from scratch using Adam optimizer (learning rate $10^{-3}$, batch 128) and categorical cross-entropy.
To prevent overfitting, we employ $\ell_{2}$ regularization ($10^{-2}$) and Early Stopping (patience 8). 
Robustness to hand laterality is improved via horizontal flipping augmentation. For fair comparison, all models are evaluated using identical stratified folds.

\paragraph{\textbf{Fusion Strategy Selection}}
\label{sec:ablation_strategy}

We evaluate four fusion depths within the same lightweight \gls{cnn} backbone under an iso-activation constraint (i.e., a fixed activation budget across layers), in order to avoid architectures that would require excessive on-chip RAM due to large intermediate tensors.
To satisfy this constraint, all variants share a three-layer convolutional backbone (\textbf{8, 16, and 32} filters), \textbf{Global Average Pooling}, and a $1\times 7$ \textbf{linear classifier}.
Fusion depth is controlled via \textbf{grouped convolutions}, isolating \gls{ir} and \gls{tof} processing streams until ungrouped layers enable cross-modal mixing. We compare the following configurations:
Under this unified backbone, we compare the following fusion-depth configurations:

\begin{itemize}
    \item \textbf{Vanilla Fusion}: Modalities are concatenated at the input and processed by standard convolutions, enabling free mixing from the first receptive field.
    \item \textbf{Early Fusion}: The first convolution uses grouped convolutions, promoting modality-specific primitive extraction before feature mixing in subsequent layers.
    \item \textbf{Mid Fusion}: The grouped constraint is extended to the second layer, postponing cross-modal mixing until higher-level representations are formed.
    \item \textbf{Late Fusion}: Separate modality-specific branches are preserved throughout feature extraction and merged only at the final classification stage.
\end{itemize}

Model selection is based on the average cross-validation accuracy.
As expected, increasing the number of grouped layers also reduces the number of learnable parameters, since grouped convolutions restrict cross-channel connectivity; deeper \textit{virtual branching} therefore yields a lighter model while progressively delaying inter-modality mixing.
Figure \ref{fig:network} shows the \textit{Early Fusion} network schematics.

\paragraph{\textbf{Multimodal Validation}}
\label{sec:eval_protocol}

After selecting the optimal fusion architecture, we quantified its benefit against single-sensor baselines to isolate the contribution of each modality.
Three configurations are compared:
\textbf{IR-only}, using the AMG8833 thermal frames;
\textbf{\gls{tof}-only}, using the VL53L8CH depth maps; and
\textbf{Combined}, using the optimal fusion architecture fed with the stacked $8\times8\times2$ multimodal tensor.

Multimodal gains are reported in terms of mean accuracy and stability across folds.
Beyond predictive performance, we also evaluate the discriminative quality of the learned embeddings by analyzing the latent feature space with \gls{tsne} visualizations and clustering indices (Silhouette Score and Davies--Bouldin Index), to determine whether fusion yields more separable class representations than single-modality models.

\paragraph{\textbf{Embedded Feasibility}}
\label{sec:embedded_methods}

To assess real-world deployability, we benchmarked inference latency on two commercial \glspl{mcu}: an \textbf{STM32F4} (Arm Cortex-M4, \SI{84}{MHz}, \SI{512}{kB} Flash, \SI{96}{kB} RAM) as a low-power baseline, and an \textbf{STM32H7} (Arm Cortex-M7, \SI{480}{MHz}, \SI{2}{MB} Flash, \SI{1}{MB} RAM) as a high-performance alternative.
All trained Keras networks were converted into embedded C for on-target execution using STMicroelectronics \textbf{Edge AI} (v3.0.0) and post-training quantized to 8-bit integers for both weights and activations to reduce memory usage and exploit Cortex-M DSP capabilities.
Power consumption was measured separately using a digital oscilloscope (i.e, Tektronik MSO64B) by monitoring the voltage drop across a \SI{100}{\milli\ohm} shunt resistor in series with the \gls{mcu} power rail, with a GPIO signal used to window the inference pass over 1,000 cycles for reproducibility \cite{Bartoli_2025}.


\vspace{-1mm}

\section{Experimental Results}
\label{sec:results}

\subsection{Fusion Strategy Selection}
Following the methodology defined in Sec.~\ref{sec:ablation_strategy}, we first evaluated the impact of fusion depth on classification performance to select the optimal architecture.

Results show that the \textbf{Early Fusion} strategy achieves the highest mean accuracy across the 5-fold validation.
In contrast, \textbf{Late Fusion} attains lower accuracy than the early-integration configurations.
In terms of complexity, increasing the number of grouped layers progressively reduces the number of learnable parameters (from 6,415 in Vanilla to 3,463 in Late Fusion).
Table~\ref{tab:ablation} reports the results of the iso-activation comparison.

\begin{table}[!t]
    \centering
    \caption{Fusion Strategy Selection (Stratified 5-Fold Cross-Validation). Comparison of fusion depths under iso-activation constraints.}
    \label{tab:ablation}
    \begin{tabular}{lccc}
        \toprule
        \textbf{Strategy} & \textbf{Groups} & \textbf{Params} & \textbf{Accuracy (Mean $\pm$ Std) [\%]} \\
        \midrule
        Vanilla Fusion          & [1, 1, 1] & 6,415 & 90.82 $\pm$ 0.41 \\
        \textbf{Early Fusion}   & [2, 1, 1] & 6,343 & \textbf{92.29 $\pm$ 1.06} \\
        Mid Fusion              & [2, 2, 1] & 5,767 & 89.80 $\pm$ 1.71 \\
        Late Fusion             & [2, 2, 2] & \textbf{3,463} & 87.04 $\pm$ 1.03 \\
        \bottomrule
    \end{tabular}
\end{table}

\subsection{Multimodal Validation}
\label{sec:results_multimodal}

Following the methodology defined in Sec.~\ref{sec:eval_protocol}, we benchmark the selected \textbf{Early Fusion} model against the single-sensor baselines to quantify the net benefit of multimodal fusion.
Table~\ref{tab:results_multimodal} reports the accuracy and stability (Mean $\pm$ Std) obtained via the stratified 5-fold cross-validation protocol.

The \textbf{Early Fusion} model achieves a mean accuracy of \textbf{92.29\%}, outperforming both the \gls{ir}-only baseline ($86.57\%$) and the \gls{tof}-only baseline ($88.65\%$) across all folds, while also achieving the lowest standard deviation ($\pm 1.06\%$).


\begin{table}[!t]
    \centering
    \caption{Multimodal validation (Stratified 5-Fold Cross-Validation). Accuracy reported as Mean $\pm$ Std over 5 folds.}
    \label{tab:results_multimodal}
    \footnotesize
    \setlength{\tabcolsep}{5pt}
    \begin{tabular}{lccc}
        \toprule
        \textbf{Model} & \textbf{Input} & \textbf{Params} & \textbf{Accuracy [\%]} \\
        \midrule
        IR Only       & $8\times8\times1$ & 6,343 & 86.57 $\pm$ 1.65 \\
        \gls{tof} Only      & $8\times8\times1$ & 6,343 & 88.65 $\pm$ 1.60 \\
        \textbf{Early Fusion} & $\mathbf{8\times8\times2}$ & \textbf{6,343} & \textbf{92.29 $\pm$ 1.06} \\
        \bottomrule
    \end{tabular}
\end{table}

\paragraph{Qualitative Analysis}
\label{par:results_qualitative}

\begin{figure*}[!t]
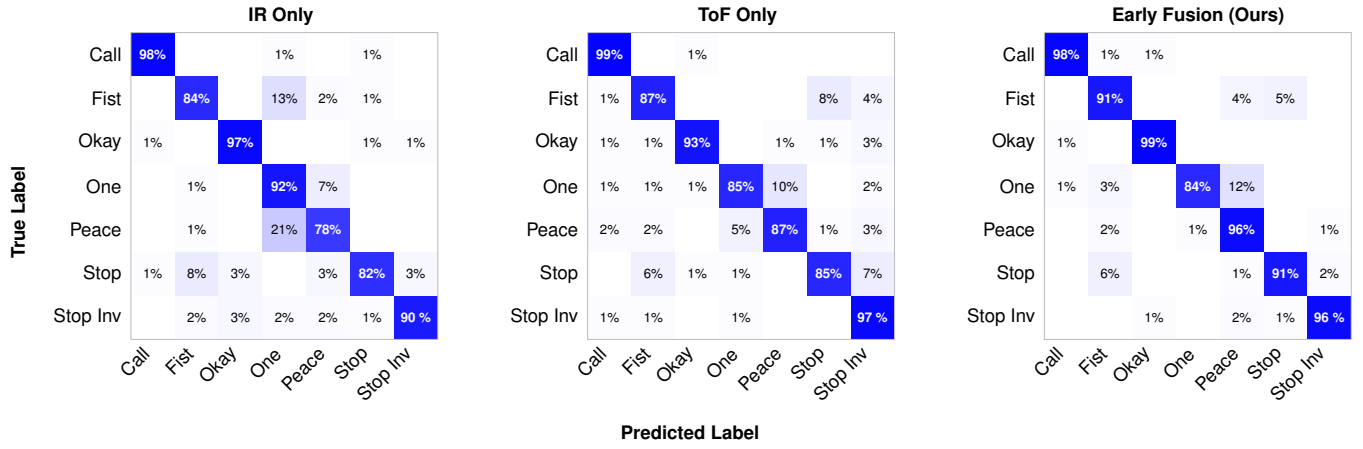

    \centering
    \setlength{\tabcolsep}{0pt}

    \begin{minipage}{0.02\textwidth}
        \centering
        \rotatebox{90}{\sffamily\scriptsize\textbf{True Label}}
    \end{minipage}%
    \begin{minipage}{0.98\textwidth}
        \centering

        \begin{minipage}[t]{0.32\textwidth}
            \centering
            {\hspace*{9mm}\sffamily\scriptsize\textbf{IR Only}\par\vspace{1mm}}
            \drawconfusionmatrix{
                98, 0, 0, 1, 0, 1, 0,
                0, 84, 0, 13, 2, 1, 0,
                1, 0, 97, 0, 0, 1, 1,
                0, 1, 0, 92, 7, 0, 0,
                0, 1, 0, 21, 78, 0, 0,
                1, 8, 3, 0, 3, 82, 3,
                0, 2, 3, 2, 2, 1, 90
            }
        \end{minipage}
        \hfill
        \begin{minipage}[t]{0.32\textwidth}
            \centering
            {\hspace*{9mm}\sffamily\scriptsize\textbf{ToF Only}\par\vspace{1mm}}
            \drawconfusionmatrix{
                99, 0, 1, 0, 0, 0, 0,
                1, 87, 0, 0, 0, 8, 4,
                1, 1, 93, 0, 1, 1, 3,
                1, 1, 1, 85, 10, 0, 2,
                2, 2, 0, 5, 87, 1, 3,
                0, 6, 1, 1, 0, 85, 7,
                1, 1, 0, 1, 0, 0, 97
            }
        \end{minipage}
        \hfill
        \begin{minipage}[t]{0.32\textwidth}
            \centering
            {\hspace*{10mm}\sffamily\scriptsize\textbf{Early Fusion (Ours)}\par\vspace{1mm}}
            \drawconfusionmatrix{
                98, 1, 1, 0, 0, 0, 0,
                0, 91, 0, 0, 4, 5, 0,
                1, 0, 99, 0, 0, 0, 0,
                1, 3, 0, 84, 12, 0, 0,
                0, 2, 0, 1, 96, 0, 1,
                0, 6, 0, 0, 1, 91, 2,
                0, 0, 1, 0, 2, 1, 96
            }
        \end{minipage}

    \end{minipage}

    \vspace{-0.1cm}
    \begin{center}
        \sffamily\scriptsize\textbf{Predicted Label}
    \end{center}
    \vspace{-0.2cm}

    \caption{Confusion matrices on the test set. Values are reported in percentages (\%). The global layout highlights the superior disambiguation capability of the Early Fusion (right) compared to \gls{ir} (left) and \gls{tof} (center), particularly for the 'Stop Inv' class.}
    \label{fig:confusion_matrices_final}
\end{figure*}

To detail the performance distribution across classes, we analyze the model corresponding to the best-performing fold.
Figure~\ref{fig:confusion_matrices_final} displays the confusion matrices for the IR-only, \gls{tof}-only, and Early Fusion models, showing the distribution of correct predictions (diagonal) and misclassifications (off-diagonal).
Table~\ref{tab:per_class} lists the corresponding F1-scores for each class.
The F1-scores for the IR-only configuration range between 0.80 and 0.98, while the \gls{tof}-only configuration yields values between 0.87 and 0.96. 
The results for the Early Fusion model are reported in the third column, with values ranging from 0.89 to 0.99.

\begin{table}[!t]
    \centering
    \caption{Per-class F1-score by modality on the test set. Best values are highlighted in bold.}
    \label{tab:per_class}
    \footnotesize
    \setlength{\tabcolsep}{4pt}
    \renewcommand{\arraystretch}{1.05}
    \begin{tabular}{lccc}
        \toprule
        \textbf{Class} & \textbf{IR Only} & \textbf{ToF Only} & \textbf{Early Fusion (Ours)} \\
        \midrule
        Call     & 0.98 & 0.96 & \textbf{0.99} \\
        Fist     & 0.86 & \textbf{0.89}& \textbf{0.89} \\
        Okay     & 0.95 & 0.95 & \textbf{0.98} \\
        One      & 0.80 & 0.88 & \textbf{0.90} \\
        Peace    & 0.81 & 0.88 & \textbf{0.89} \\
        Stop     & 0.88 & 0.87 & \textbf{0.92} \\
        Stop Inv & 0.92 & 0.90 & \textbf{0.96} \\
        \bottomrule
    \end{tabular}
\end{table}

\begin{figure*}[t!]
    \centering
    \includegraphics[width=.9\textwidth]{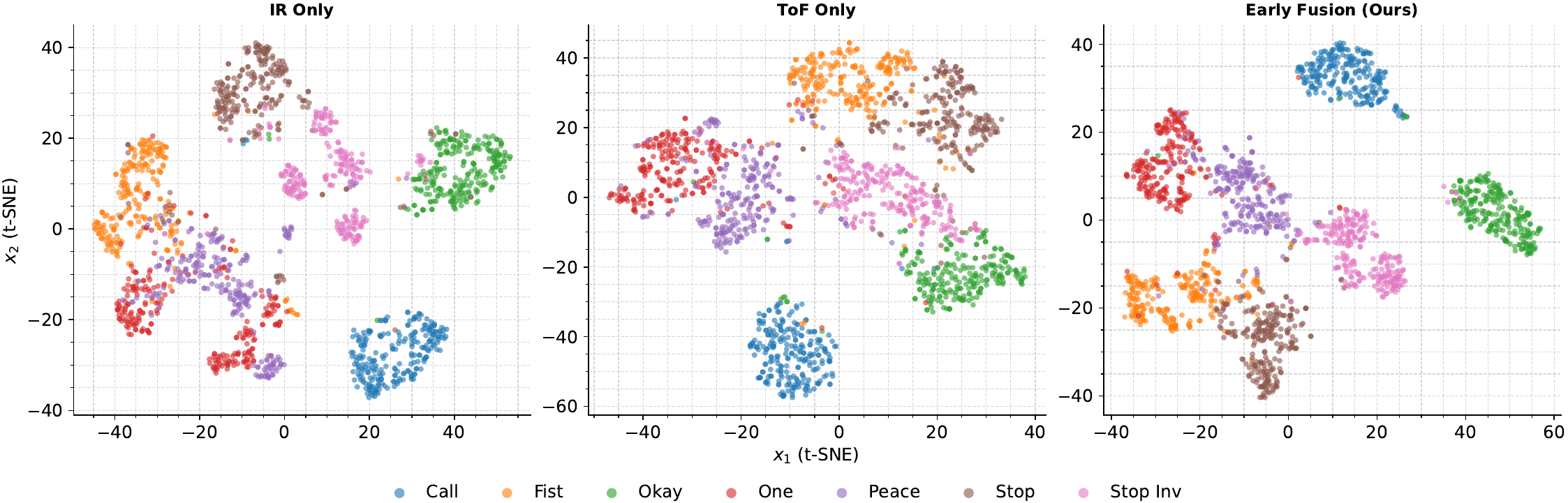}
    \vspace{-0.2cm}
    \caption{Latent space visualization using t-SNE projections of the test set embeddings. The plots display the feature distribution for IR-only (left), ToF-only (center), and Early Fusion (right) configurations.}
    \label{fig:tsne_analysis}
\end{figure*}

\paragraph{Latent Space Quality}
Finally, we analyze the geometric structure of the learned representations. 
Figure~\ref{fig:tsne_analysis} visualizes the \gls{tsne} projection of the 32-dimensional feature embeddings extracted from the penultimate layer of the networks.
Table~\ref{tab:cluster_metrics} reports the cluster quality metrics computed directly on the high-dimensional feature space.
The IR-only model yields a Silhouette Score of 0.248 and a Davies-Bouldin Index of 1.395, while the ToF-only model results in values of 0.234 and 1.580, respectively.
The Early Fusion model records a Silhouette Score of \textbf{0.301} and a Davies-Bouldin Index of \textbf{1.283}.

\begin{table}[!t]
    \centering
    \caption{Cluster Quality Metrics computed on the 32D latent vectors. Arrows indicate the direction of better performance.}
    \label{tab:cluster_metrics}
    \begin{tabular}{lcc}
        \toprule
        \textbf{Modality} & \textbf{Silhouette Score} ($\uparrow$) & \textbf{Davies-Bouldin} ($\downarrow$) \\
        \midrule
        IR Only & 0.248 & 1.395 \\
        \gls{tof} Only & 0.234 & 1.580 \\
        \textbf{Early Fusion} & \textbf{0.301} & \textbf{1.283} \\
        \bottomrule
    \end{tabular}
\end{table}

\subsection{Embedded Feasibility}
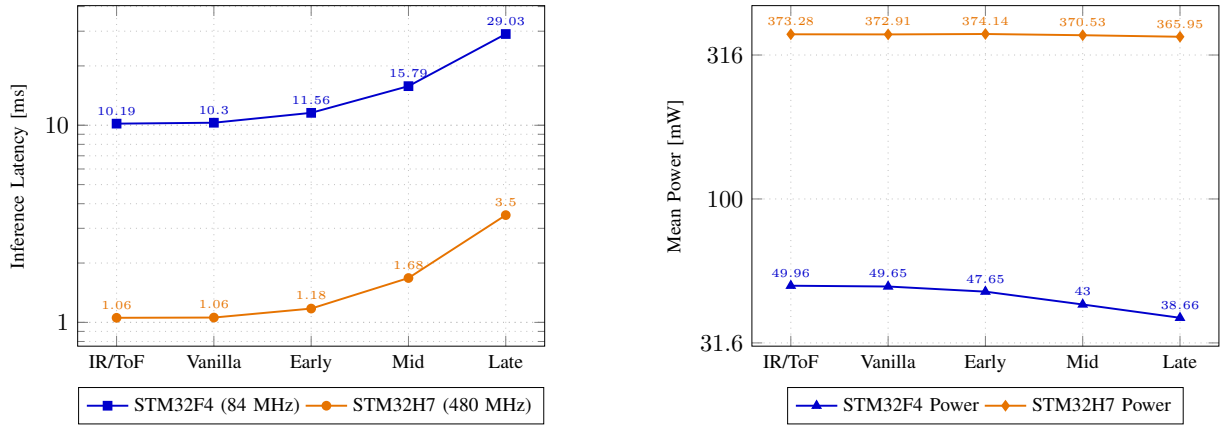
\begin{figure*}[!t]
    \centering
    \resizebox{0.9\linewidth}{!}{
    \begin{tikzpicture}
    \begin{groupplot}[
        group style={
            group size=2 by 1,
            horizontal sep=3cm
        },
        width=0.46\textwidth,
        height=6.5cm,
        symbolic x coords={IR/ToF, Vanilla, Early, Mid, Late},
        xtick=data,
        xticklabel style={font=\footnotesize},
        ylabel style={font=\footnotesize},
        grid=both,
        grid style={dotted, gray!50},
        legend style={
            at={(0.5,-0.12)},
            anchor=north,
            legend columns=-1,
            font=\footnotesize
        },
        nodes near coords,
        nodes near coords style={
            font=\tiny,
            anchor=south,
            /pgf/number format/fixed,
            /pgf/number format/precision=2
        },
        point meta=explicit
    ]
    \nextgroupplot[
        ylabel={Inference Latency [ms]},
        ymode=log,
        log ticks with fixed point,
        yticklabel style={
            /pgf/number format/fixed,
            /pgf/number format/precision=1
        }
    ]
    \addplot[
        color=blue!80!black,
        mark=square*,
        thick,
        solid,
        mark options={scale=0.8}
    ]
    coordinates {
        (IR/ToF, 10.19) [10.19]
        (Vanilla, 10.30) [10.30]
        (Early, 11.56) [11.56]
        (Mid, 15.79) [15.79]
        (Late, 29.03) [29.03]
    };
    \addlegendentry{STM32F4 (84 MHz)}
    \addplot[
        color=orange!90!black,
        mark=*,
        thick,
        solid,
        mark options={scale=0.8}
    ]
    coordinates {
        (IR/ToF, 1.055) [1.06]
        (Vanilla, 1.058) [1.06]
        (Early, 1.175) [1.18]
        (Mid, 1.679) [1.68]
        (Late, 3.501) [3.50]
    };
    \addlegendentry{STM32H7 (480 MHz)}
    \nextgroupplot[
        ylabel={Mean Power [mW]},
        ymode=log,
        log ticks with fixed point,
        yticklabel style={
            /pgf/number format/fixed,
            /pgf/number format/precision=1
        }
    ]
    \addplot[
        color=blue!80!black,
        mark=triangle*,
        thick,
        solid,
        mark options={scale=0.9}
    ]
    coordinates {
        (IR/ToF, 49.96) [49.96]
        (Vanilla, 49.65) [49.65]
        (Early, 47.65) [47.65]
        (Mid, 43.00) [43.00]
        (Late, 38.66) [38.66]
    };
    \addlegendentry{STM32F4 Power}
    \addplot[
        color=orange!90!black,
        mark=diamond*,
        thick,
        solid,
        mark options={scale=0.9}
    ]
    coordinates {
        (IR/ToF, 373.28) [373.28]
        (Vanilla, 372.91) [372.91]
        (Early, 374.14) [374.14]
        (Mid, 370.53) [370.53]
        (Late, 365.95) [365.95]
    };
    \addlegendentry{STM32H7 Power}
    \end{groupplot}
    \end{tikzpicture}
    }
    \caption{On-device inference latency (left) and mean active power (right) across fusion strategies, both shown with logarithmic y-axes.}    \label{fig:latency_power_grouped}
\end{figure*}

The proposed Early Fusion model requires only 6,343 parameters.
In an 8-bit quantized format, the model weights occupy roughly \SI{7}{kB} of Flash memory, fitting comfortably within ultra-low-power \gls{mcu}s.
While all fusion strategies were designed under a strict iso-activation constraint to maintain a constant peak RAM usage across architectures, Figure~\ref{fig:latency_power_grouped} reveals a divergent behavior in execution time and power consumption. 
On the STM32F4, the \textbf{Early Fusion} model achieves an inference latency of \SI{11.56}{ms} with an average active power of \SI{47.65}{mW}. 
Despite having fewer parameters than the Early Fusion model, the deeper split-branch strategies (Mid and Late Fusion) exhibit significantly higher inference latencies on both the STM32F4 and STM32H7 platforms.

\section{Discussion}
\label{sec:discussion}

The results confirm that fusing low-resolution thermal and depth signals improves gesture-recognition robustness while successfully avoiding the significant deployment bottlenecks of state-of-the-art alternatives. 
Unlike depth-based ResNets \cite{Ma_2024} that demand massive SRAM, or thermal-based solutions relying on theoretical FLOPS \cite{Vandersteegen_2020} or custom neuromorphic hardware \cite{Safa_2024}, our multimodal system operates efficiently on standard \glspl{mcu}. 

Regarding the architectural design, our analysis reveals a direct trade-off between accuracy and efficiency.
Under iso-activation constraints, fusion depth is inversely related to model capacity: deeper strategies such as \textbf{Late Fusion} isolate modalities into independent branches, reducing parameters from 6.4k to 3.5k. 
This compactness, however, reduces representational power, leading to an accuracy drop to 87.0\% and indicating that fully separated branches struggle to learn rich features without increasing the filter budget.

Counter-intuitively, despite having the smallest footprint, Late Fusion is the slowest, with an approximately \(3\times\) latency increase. This result suggests that a lower parameter count does not necessarily translate into faster execution on embedded targets when computation is distributed across separate branches.
Conversely, our proposed \textbf{Early Fusion} strategy strikes the optimal balance: it introduces a negligible latency penalty ($\approx 11\text{--}13\%$) compared to single-sensor baselines while maintaining the same Flash footprint, effectively delivering the best performance profile.
Prioritizing footprint over latency, Mid Fusion offers an $\approx 9\%$ parameter reduction while still outperforming unimodal baselines (89.80\% vs.\ 88.65\% ToF, 86.57\% IR).

The Combined model actively resolves ambiguities by synthesizing complementary cues.
\gls{ir} thermal gradients disambiguate hand orientation (e.g., \texttt{Stop} vs. \texttt{Stop Inv}), while \gls{tof} boundaries resolve shape errors (e.g., \gls{ir} misclassifies 21\% of \texttt{Peace} as \texttt{One}).
Residual \texttt{Peace} vs. \texttt{One} confusion arises from the coarse $8\times8$ grid, which struggles to resolve the subtle phalangeal separation characterizing multi-finger gestures.
Despite this limitation, Early Fusion achieves 92.3\% accuracy, outperforming the strongest unimodal baseline (\gls{tof} at 88.6\%) with no size penalty.

This accuracy gain stems from a better-structured 32-dimensional latent space. \textbf{Early Fusion} yields denser, more separable clusters than unimodal baselines, achieving a superior Silhouette Score (0.301 vs.\ 0.248 \gls{ir}, 0.234 \gls{tof}) and a lower Davies-Bouldin Index (1.283 vs.\ 1.395 \gls{ir}, 1.580 \gls{tof}).
These metrics confirm that the multimodal feature space is more informative and separable. 
This suggests that the model has learned robust, invariant features, merging thermal gradients with geometric shapes, that are less sensitive to noise than features derived from a single physical domain.

The synergy between low-latency inference and duty-cycling is pivotal for wearable viability. 
While sensing constitutes a constant power floor ($\approx$47.2~mW), the rapid execution of \textbf{Early Fusion} ensures the \gls{mcu} remains active for less than 6\% of the 5~Hz duty cycle. 
This efficiency prevents computation from dominating the overall energy budget, keeping total system consumption at approximately \SI{50}{mW}. 
Based on a nominal 200~mAh battery (\(\approx\)\SI{740}{mWh}) and considering steady-state sensing and inference only, the resulting operating time is estimated to be close to 15~hours. 
While this back-of-the-envelope calculation does not account for all platform-level overheads, it suggests that deeper fusion strategies would be less favorable in practice because of their longer active-state occupancy.

\section{Conclusion}
\label{sec:conclusion}

This work presented a multimodal sensor fusion framework for privacy-preserving gesture recognition on resource-constrained wearable devices. By combining low-resolution thermal and depth sensing, we showed that a lightweight \gls{cnn} can achieve accurate recognition without capturing identifiable biometric information.

Among the evaluated architectures, \textbf{Early Fusion} provided the best trade-off under iso-activation constraints, outperforming both unimodal baselines with 92.3\% accuracy and superior latent-space separability. This result is achieved with fewer than 6.4k parameters and only a small latency overhead, confirming the suitability of the proposed approach for always-on inference on standard \glspl{mcu}.

A limitation of the present study is the small number of enrolled participants, which limits subject diversity and prevents a robust assessment of user-independent generalization. 
A larger and more diverse data collection campaign will enable stronger subject-wise evaluation, including leave-one-subject-out validation.
Future work will extend the framework to dynamic gesture recognition via temporal modeling to increase real-world applicability. Additionally, to further optimize the hardware deployment, we will investigate quantization-aware training as an alternative to post-training quantization, and perform broader cost-benefit analyses evaluating the end-to-end system power overhead against competing wearable architectures.

\bibliographystyle{IEEEtran}
\bibliography{bibliography}

\end{document}